\documentclass{article}
\usepackage{arxiv}

\usepackage[caption=false,font=footnotesize]{subfig}

\usepackage{times} 
\usepackage{amsmath} 
\usepackage{amssymb}  
\usepackage{color}
\usepackage{url}
\usepackage{tikz}
\usepackage{xspace}
\usepackage{comment}

\newcommand{\bptt}{{\sc bptt}\xspace}

\newcommand{\rnn}{{\sc rnn}\xspace}
\newcommand{\rnns}{{\sc rnn}s\xspace}

\newcommand{\lstm}{{\sc lstm}\xspace}

\newcommand{\rbms}{{\sc rbm}s\xspace}
\newcommand{\rbm}{{\sc rbm}\xspace}
\newcommand{\grbms}{{\sc grbm}s\xspace}
\newcommand{\grbm}{{\sc grbm}\xspace}

\newcommand{\crbm}{{\sc crbm}\xspace}
\newcommand{\crbms}{{\sc crbm}s\xspace}
\newcommand{\gae}{{\sc gae}\xspace}
\newcommand{\gaes}{{\sc gae}s\xspace}

\newcommand{\dae}{{\sc dae}\xspace}
\newcommand{\pgp}{{\sc pgp}\xspace}

\newcommand{\cnns}{{\sc cnn}s\xspace}

\newcommand{\enc}{\mathbf{h}}
\newcommand{\dec}{\mathbf{r}}
\newcommand{\x}{\mathbf{x}}
\newcommand{\y}{\mathbf{y}}
\newcommand{\z}{\mathbf{z}}
\newcommand{\vsig}{\mathbf{\sigma}}
\newcommand{\style}{\mathbf{s}}
\newcommand{\h}{\mathbf{h}}
\newcommand{\f}{\mathbf{f}}
\newcommand{\g}{\mathbf{g}}
\newcommand{\oo}{\mathbf{o}}
\newcommand{\p}{\mathbf{p}}
\newcommand{\q}{\mathbf{q}}

\newcommand{\bb}{\mathbf{b}}
\newcommand{\B}{\mathbf{B}}
\newcommand{\W}{\mathbf{W}}
\newcommand{\MP}{\mathbf{P}}

\newcommand\diag{\textmd{diag}} 

\newcommand{\inp}{{\mbox{\scriptsize in}}}
\newcommand{\out}{{\mbox{\scriptsize out}}}

\usepackage{multicol}

\usepackage{algorithmic}

\makeatletter
\newcounter{algorithmbis}
\renewcommand{\thealgorithmbis}{\thesection.\arabic{algorithmbis}}
\def\algorithmbis{\@ifnextchar[{\@algorithmbisa}{\@algorithmbisb}}
\def\@algorithmbisa[#1]{%
  \refstepcounter{algorithmbis}
  \trivlist
  \leftmargin\z@
  \itemindent\z@
  \labelsep\z@
  \item[\parbox{\linewidth}{%
    \hrule
    \hrule
    \noindent\strut\textbf{Algorithm \thealgorithmbis} #1
    \hrule
  }]\hfil\vskip0em%
}
\def\@algorithmbisb{\@algorithmbisa[]}

\makeatother

\definecolor{myred}{rgb}{0.8,0,0}
\definecolor{mygreen}{rgb}{0,0.6,0}
\definecolor{myblue}{rgb}{0,0,0.7}

\newcommand{\todolater}[1]{\textcolor{myblue}{}}

\renewcommand{\cite}{\citep} 

\begin{document}

\title{Gated networks: an inventory}

\author{Olivier Sigaud\\
Sorbonne Universit\'es, UPMC Univ Paris 06, UMR 7222, F-75005 Paris, France\\ 
CNRS, Institut des Syst\`emes Intelligents et de Robotique UMR7222, Paris, France\\
{\tt olivier.sigaud@isir.upmc.fr}~~~~+33 (0) 1 44 27 88 53
\And
Cl\'ement Masson, David Filliat, Freek Stulp\\
\'Ecole Nationale Sup\'erieure de Techniques Avanc\'ees (ENSTA-ParisTech) \\
FLOWERS Research Team, INRIA Bordeaux Sud-Ouest.\\ 
828, Boulevard des Mar\'echaux, 91762 Palaiseau Cedex, France\\
{\tt \{clement.masson,david.filliat,freek.stulp\}@ensta-paristech.fr}}

\maketitle


\begin{abstract}
Gated networks are networks that contain gating connections, in which the outputs of at least two neurons are multiplied. Initially, gated networks were used to learn relationships between two input sources, such as pixels from two images. More recently, they have been applied to learning activity recognition or multimodal representations. 
The aims of this paper are threefold:
1)~to explain the basic computations in gated networks to the non-expert, while adopting a standpoint that insists on their symmetric nature. 
2)~to serve as a quick reference guide to the recent literature, by providing an inventory of applications of these networks, as well as recent extensions to the basic architecture.
3)~to suggest future research directions and applications. 
\end{abstract}


\section{Introduction}

Due to its many successful applications to pattern recognition, deep learning has become one of the most active research trends in the machine learning community \cite{LeCun15nature}.
The main building blocks in the deep learning literature are Restricted Boltzmann Machines (\rbms)~\cite{smolensky1986information}, autoencoders \cite{Hinton2006,Vincent2008}, Convolutional Neural Networks (\cnns) \cite{lecun1998gradient} and Recurrent Neural Networks (\rnns) \cite{bengio2013deep}.

Most of these architectures are used to learn a relationship between a single input source and the corresponding output. They do so by building a representation of the input domain that facilitates the extraction of the adequate relationship. However, there are many domains where the representation to be learned should relate more than one source of input to the output.

In reinforcement learning, for instance, value functions take a state and an action as input, and output a expected return. In order to deal with continuous states and actions, finding separately the adequate representations for states and actions to facilitate value function learning might be critical \cite{mnih2015human,lillicrap2015continuous}. Moreover, there are cases where learning a {\em reversible} tripartite relationship might be particularly useful. For instance, in control problems, forward models take a state and an action as input, and output the next state whereas inverse models take the current state and a desired state as input, and output an action. It would be interesting to learn a single representation for both models which could be used both in the forward and the inverse way.

\begin{figure}[bht]
\centering
\includegraphics[width=0.8\textwidth]{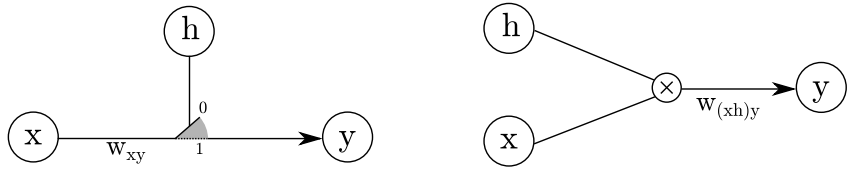}
\caption{\label{fig:gating_neurons} Two types of gating connections. On the left hand-side, the h neuron acts as a switch or gate that stops or not the flow of information between $x$ and $y$. On the right hand-side, the connection implements a multiplicative relationships between the inputs $x$ and $h$ to provide the output $y$. Image reproduced from \cite{droniouPHD}.}
\end{figure}

Gated networks are extensions of the above deep learning building blocks that are designed to learn relationships between at least two sources of input and at least one output.
A defining feature of these architectures is that they contain \emph{gating connections}, as visualized in \figurename~\ref{fig:gating_neurons}. When the relationships between several sources of data involves multiplicative interactions, such gating connections between neurons result in more natural topologies and increase the expressive power of neural networks, because implementing a multiplicative relationship between two layers of standard neurons would require a number of dedicated neurons that would grow exponentially with the required precision \cite{memisevic2013learning}.

Although the history of gating connections dates back at least to 1981 \cite{memisevic2013learning}, there has been a recent surge of interest in these networks.
Initially they were mainly used to learn transformation between images \cite{Memisevic2007}, but they have recently also been applied to human activity recognition from videos and moving skeleton data from the kinect sensor \cite{mocanu2015factored}, or to recognize offspring relationship from pictures of faces \cite{dehghan2014look}.

In robotics, gated networks have been used to learn to write numbers \cite{droniou14ICDL}, as well as to learn multimodal representations of numbers using images, vocal signal and articular movements with the iCub robot \cite{droniou14RAS}. At a higher level, the same tools could be used to learn affordances, that are often represented as object-action-effect complexes \cite{Montesano2008}. All these examples have led to the claim that gated networks might be a particularly suitable tool along the way towards {\em deep developmental robotics} \cite{sigaud2015towards}.

Despite this growing interest, the literature about gated networks is still sparse enough so that it can be covered into a short survey. The aims of this paper are to cover the basics of gated networks for the non-expert, to serve as an inventory of applications of gated networks, and to suggest future research directions and applications.

The rest of this paper is structured as followed. In the next section, we give a detailed account of the calculations performed by the standard gated network model and a few variants whose relationship to the standard model is highlighted after some generalization. In this presentation, we emphasize the symmetric nature of these networks because it reveals the connections between some of the surveyed works. Then we present the standard unsupervised learning mechanisms that are used for tuning these networks, and provide an inventory of the various uses which is summarized into a table. Finally, we survey a few recent architectures that include the core ingredient of gated networks, and conclude with directions for future research. 

\section{Standard Gated Network Architectures}

Gated networks are networks where the input of some computational units (or ``neurons'') is a function of the product of the output of several other neurons.
As illustrated in \figurename~\ref{fig:gating_neurons}, one can consider two kinds of connections between 3 neurons.
In the first family, a neuron $h$ is used as a switch that stops or not the flow of information between two other neurons $x$ and $y$. This functionality is very similar to that of transistors as electronic switches in digital circuits. This mechanism is used in the \lstm family of networks \cite{hochreiter1997lstm,salak15}, among others. In the second family, the gating connection implements a multiplicative relationship between two inputs $x$ and $y$. Note that the latter mechanism is more general than the former, since a value of 0 in $h$ gates $y$ to 0. The most general view is that neuron $h$ modulates the signal between $x$ and $y$.

In this paper, we focus on the specific family of neural networks implementing a multiplicative relationship that are built on \rbms and autoencoders and, to a lesser extent, on \cnns and \rnns.

\subsection{From gated \rbms to gated autoencoders and beyond}
\label{sec:history}

We now briefly introduce Restricted Boltzmann Machines (\rbms)~\cite{smolensky1986information}, autoencoders \cite{Hinton2006,Vincent2008}, Convolutional Neural Networks (\cnns) \cite{lecun1998gradient} and Recurrent Neural Networks (\rnns) \cite{bengio2013deep}, and show how these networks have been extended to contain gated connections.

An \rbm is not a neural network but a particular probabilistic graphical model (PGM) \cite{koller2009probabilistic} whose graph is bipartite: one set (or layer) of nodes is called ``visible'' and is used as the input of the model, whereas the other layer is ``hidden'' and is interpreted as being the hidden cause explaining the input. Both layers are generally binary (though it is possible to extend them to real-valued units) and fully connected to each other. However,  there are no connections within a layer, which facilitates inference and training. Training an \rbm consists in finding the parameters (edge's weights and node's bias) which maximize the likelihood of the training data.
Importantly, \rbms are {\em generative} models: they can model the probability density of the joint distribution of visible and hidden units, which enables them to generate samples similar to those of the training data onto the visible layer.

The first instance of a gated network in the deep learning literature was a gated \rbm (\grbm) \cite{Memisevic2007}. However, this model was using a fully connected multiplicative network that required a lot of memory and computations for inference and training. In the next section, we present a solution to this issue, that was introduced by \citet{Memisevic2010} as a direct extension of \cite{Memisevic2007}, still using \grbms.

Autoencoders also contain an input and a representation layer but, in contrast to \rbms, they are deterministic models. They are trained to encode the input into the latent representation layer and then to reconstruct (or decode) the input from that representation. In their basic form, they are {\em discriminative} models, which can only compute the hidden layer given an input.
It was then shown that a particular class of regularized autoencoder, the denoising autoencoder (\dae), could learn a model of the data generating distribution.
This endow autoencoders with generative properties similar to those of \rbms \cite{Vincent2008}. More formally, a \dae can be interpreted as a Gaussian \rbm \cite{vincent11}.

This led to a shift from \grbms to gated autoencoders (\gaes) \cite{memisevicPHD,memisevic2011gradient,Memisevic2012a} though research on \grbms is still active \cite{Taylor2010,ding2014mental}.

Convolutional Neural Networks are an early family of deep learning architectures which are composed of alternating convolutional layers and pooling layers.
They are inspired from the human vision system and they proved particularly efficient for image processing applications. Finally, \rnns contain at least one recurrent connection, which makes them adequate for dealing with temporally extended information \cite{schmidhuber1997lstm}.

The gating idea was also applied to \rnns \cite{sutskever2011generating} and \cnns, either combined to \grbms \cite{Taylor2010} or to \gaes \cite{konda2015learning}, as we outline in Section~\ref{sec:beyond}.

\subsection{Reducing the number of multiplicative connections}
\label{sec:trick}

Implementing a gated network requires memory. Consider the network shown in \figurename~\ref{fig:gsc}(a), consisting of three layers $\x$, $\y$ and $\h$\footnote{Throughout this document, bold lowercase symbols denote vectors, and bold uppercase symbols denote matrices.} whose respective cardinality is $n_x$, $n_y$ and $n_h$. 
Predicting the output layer $\hat{\y}$ given $\x$ and $\h$ with such a multiplicative network consists in computing all the values $\hat{y}_j$ of $\hat{\y}$ using
\begin{equation}
\label{eq:3waytensor}
\forall j, \hat{y}_j = \sigma_y(\sum_{i = 1}^{n_x}\sum_{k = 1}^{n_h} W_{ijk} x_i h_k)
\end{equation}
where $\sigma_y$ is some (optional) non-linear {\em activation} function described in more details in Section~\ref{sec:activ}.

Alternatively, one may compute $\hat{\x}$ given $\y$ and $\h$ or compute $\hat{\h}$ given $\x$ and $\y$ using

$$\forall i, \hat{x_i} = \sigma_x(\sum_{j = 1}^{n_y}\sum_{k = 1}^{n_h} W_{ijk} y_j h_k), \hspace{0.8cm} \forall k, \hat{h_k} = \sigma_h(\sum_{i = 1}^{n_x}\sum_{j = 1}^{n_y} W_{ijk} x_i y_j).$$

Regardless of the $\sigma$ functions, these models are called {\em bilinear} because, if one input is held fixed, the output is linear in the other input.

The weights $W_{ijk}$ define a \emph{3-way tensor}, which is used to compute $\hat{\x}$, $\hat{\y}$ or $\hat{\h}$ given both other vectors. This tensor contains $n_x \times n_y \times n_h$ connections. If $n_x$, $n_y$ and $n_h$ are in the same order of magnitude, the number of weights (aka parameters) is cubic in this magnitude.

Factored architectures are designed to avoid representing this cubic number of weights. Two ways to reduce this memory requirement are:
\begin{itemize}
\item
Projecting the input and output, potentially high-dimensional signals, into a smaller space through \emph{factor layers}, and then performing the central product between these smaller dimensions.
\item
Constraining the structure of the global 3-way tensor so as to restrict the number of weights.
\end{itemize}

In the next two sections, we show that the standard gated network takes the best of both views, by setting a constraint on the 3-way tensor structure that implements a projection onto factor layers, but that also avoids representing the full central product. Another striking feature of this architecture is that the resulting central product does not contain any tunable parameter. 

\subsubsection{Projecting onto factor layers}
\label{sec:project}

\begin{figure}[htb]
\centering
\includegraphics[width=0.95\textwidth]{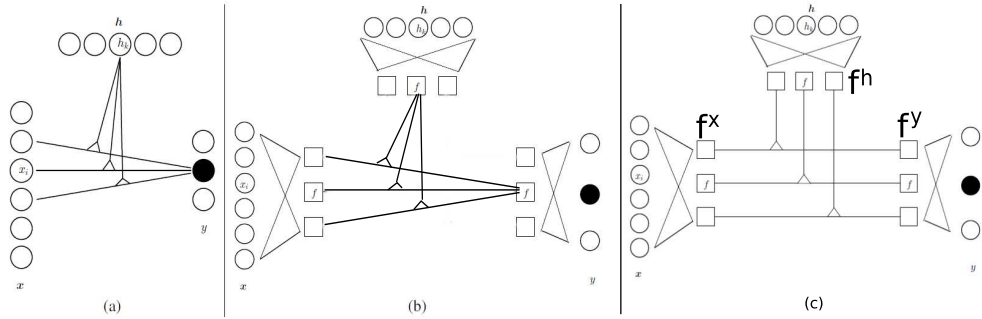}
\caption{(a): A fully connected multiplicative network. (b): A simplified network introducing factor layers. (c): The factored gated architecture.
All figures are adapted from \cite{Memisevic2010b}.
}
\label{fig:gsc}
\end{figure}

One way of reducing the number of weights consists in projecting the $\x$, $\y$ and $\h$ layers onto smaller layers noted respectively $\f^x$, $\f^y$ and $\f^h$ before performing the product between these smaller layers. Given their multiplicative role, these layers are called ``factor'' layers.
The corresponding approach is illustrated in \figurename~\ref{fig:gsc}(b).
If the respective cardinality of the factors is $n_{f_x}$, $n_{f_y}$ and $n_{f_h}$, the number of weights of the central 3-way tensor is $n_{f_x} \times n_{f_y} \times n_{f_h}$.
To tune the whole network, additional weights must be added to this 3-way tensor, respectively $n_x \times n_{f_x}$, $n_y \times n_{f_y}$ and $n_h \times n_{f_h}$ for each layer, so the total number of weights is $(n_{f_x} \times n_{f_y} \times n_{f_h}) + (n_x \times n_{f_x}) + (n_y \times n_{f_y}) + (n_h \times n_{f_h})$.

In summary, the two input layers among $\x$, $\y$ and $\h$ are first projected onto feature spaces through the corresponding factor layers, then the central 3-way multiplication is performed and finally projected to the output layer through the last factor layer. 

This approach, suggested by \citet{Memisevic2007}, results in fewer tunable parameters provided that factor layers contain fewer neurons than the input layers.
In that case, the network performs {\em dimensionality reduction} on the inputs before tuning the multiplicative weights between the factors. As a result, the number of weights is still cubic, but of a smaller magnitude. A second benefit of this architecture is that, in contrast to the one illustrated in \figurename~\ref{fig:gsc}(a), the introduction of factors results in the possibility of {\em feature sharing} between the different external layers \cite{Memisevic2010b}. However, to the best of our knowledge, this way of reducing the number of parameters of the gated architecture has not yet been implemented.

Another approach, which is used in all the works surveyed hereafter, consists in rather calling upon {\em over-complete} representations \cite{olshausen2003principles}, where
factor layers are larger than the input space, but a regularization method like {\em denoising} \cite{vincent11} is used to sparsify the activity of the factors. In this context, introducing the factor layers does not reduce the number of parameters, it even increases it \cite{memisevic2013learning}.

\subsubsection{Constraining the 3-way tensor}
\label{sec:constrain}

Another way of reducing the number of parameters consists in restricting the weights $W_{ijk}$ to follow a specific form
\begin{equation}
\label{eq:factors}
W_{ijk} = \sum_{f = 1}^F W^x_{if} W^y_{jf} W^h_{kf}.
\end{equation}

With this constraint, the matrices $\W^x$, $\W^y$ and $\W^h$ are of respective size $n_x \times n_f$, $n_y \times n_f$ and $n_h \times n_f$, thus the total number of weights is just $n_f\times(n_x+n_y+n_z)$, which is quadratic instead of cubic in the size of input or factors.

Consider again the case where $\hat{\y}$ is predicted given $\x$ and $\h$. Equation (\ref{eq:3waytensor}) can be rewritten as
\begin{equation}
\label{eq:t1}
\forall j, \hat{y}_j = \sigma_y(\sum_{i = 1}^{n_x}\sum_{k = 1}^{n_h} \sum_{f = 1}^F W^x_{if} W^y_{jf} W^h_{kf} x_i h_k),
\end{equation}
which can be reorganized into
\begin{equation}
\label{eq:t2}
\forall j, \hat{y}_j = \sigma_y(\sum_{f = 1}^F W^y_{jf}( \sum_{i = 1}^{n_x} W^x_{if} x_i)(\sum_{k = 1}^{n_h} W^h_{kf} h_k)).
\end{equation}

By noting 
\begin{equation}
\label{eq:notations}
f^x_f = \sum_{i = 1}^{n_x} W^x_{if} x_i, \hspace{0.5cm} f^y_f = (\sum_{j = 1}^{n_y} W^y_{jf} y_j), \hspace{0.5cm} f^h_f = (\sum_{k = 1}^{n_h} W^h_{kf} h_k),
\end{equation}

we finally get
\begin{equation}
\label{eq:t3}
\forall j, \hat{y}_j = \sigma_y(\sum_{f = 1}^F W^y_{jf} f^x_f.f^h_f).
\end{equation}
The three equations in (\ref{eq:notations}) define three factor layers as explained in Section~\ref{sec:project} and illustrated in \figurename~\ref{fig:gsc}(b).
However, when looking at the structure of (\ref{eq:t3}), one can see that, instead of having a full central product, the output of both factor layers --~$\f^x$ and $\f^h$ in the case of (\ref{eq:t3})~-- are multiplied element-wise through the same index $f$, as illustrated in \figurename~\ref{fig:gsc}(c).

Thus, using the decomposition of (\ref{eq:t3}), it can be seen that this way of constraining the 3-way tensor corresponds to using projections as in the previous view, but with three factor layers $\f^x$, $\f^y$ and $\f^h$ of the same size $n_f$, and where the central 3-way tensor has been replaced by 3 two-by-two element-wise products of the factor layers.

With a more algebraic notation,  (\ref{eq:notations}) can be rewritten
\begin{equation}
\label{eq:algebra}
\f^x = {\W^x}^\intercal \x, \hspace{0.5cm} \f^y = {\W^y}^\intercal \y, \hspace{0.5cm} \f^h = {\W^h}^\intercal \h.
\end{equation}

\begin{figure}[htp]
\centering
\includegraphics[width=0.9\textwidth]{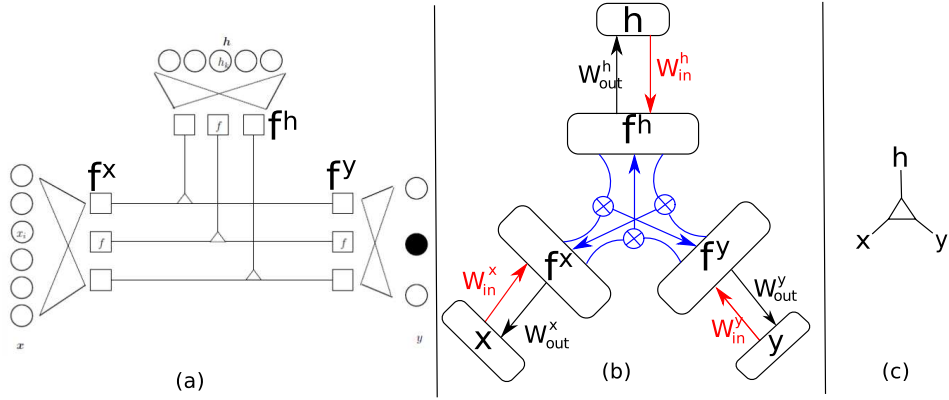}
\caption{Two views of \gaes. (a): Same as \figurename~\ref{fig:gsc}(c). (b): Another view of the same architecture, adapted from \cite{droniouPHD}. (c): Simplified notation corresponding to (b), reused in figures \ref{fig:taylor}, \ref{fig:droniouRAS} and \ref{fig:sutskever}.
}
\label{fig:gated_connections}
\end{figure}

In this notation, we omit the representation of an additive bias term by considering the inputs as being a homogeneous representation with an additional constant value, in which biases are implemented implicitly. Equation (\ref{eq:t3}) then becomes
\begin{equation}
\label{eq:alg_yxh}
\hat{\y} = \vsig_y(\W^y (\f^x \otimes \f^h)),
\end{equation}
where $\otimes$ denotes the element-wise multiplication illustrated in \figurename~\ref{fig:gated_connections}(b).

Again, the same decomposition can be applied to predict $\hat{\h}$ given $\x$ and $\y$ or to predict $\hat{\x}$ given $\y$ and $\h$, giving rise to 
\begin{equation}
\label{eq:alg_xh}
\hat{\x} = \vsig_x(\W^x (\f^y \otimes \f^h)), \hspace{0.5cm} \hat{\h} = \vsig_h(\W^h (\f^x \otimes \f^y)).
\end{equation}

A slightly more general version of the same architecture that insists on its symmetric nature can be obtained by noting $\W^x_\inp$, $\W^y_\inp$ and $\W^h_\inp$ the matrices oriented from the input layers towards the factors, and $\W^x_\out$, $\W^y_\out$ and $\W^h_\out$ those oriented from the factors towards the output. The corresponding architecture is depicted in \figurename~\ref{fig:gated_connections}(b).

Following these notations, if we consider computations from the input layers to the factors, the red arrows correspond to

$$\f^x_\inp = \W^x_\inp \x, \hspace{0.5cm} \f^y_\inp = \W^y_\inp \y, \hspace{0.5cm} \f^h_\inp = \W^h_\inp \h.$$

In the other way, from the factors to other factors, the blue arrows correspond to

$$\f^x_\out = \f^y_\inp \otimes \f^h_\inp, \hspace{0.5cm} \f^y_\out = \f^x_\inp \otimes \f^h_\inp, \hspace{0.5cm} \f^h_\out = \f^x_\inp \otimes \f^y_\inp. $$

Finally, towards the output we have

$$\hat{\x} = \vsig_x(\W^x_\out \f^x_\out), \hspace{0.5cm} \hat{\y} = \vsig_y(\W^y_\out \f^y_\out), \hspace{0.5cm} \hat{\h} = \vsig_h(\W^h_\out \f^h_\out).$$

By connecting the above elements together, the complete input-output functions are
\begin{equation}
\label{eq:hxy}
\hat{\h} = \oo(\x, \y) = \vsig_h(\W^h_\out((\W^x_\inp \x) \otimes (\W^y_\inp \y))),
\end{equation}
\begin{equation}
\label{eq:xyh}
\hat{\x} = \p(\y, \h) = \vsig_x(\W^x_\out ((\W^y_\inp \y) \otimes (\W^h_\inp \h))),
\end{equation}
\begin{equation}
\label{eq:yxh}
\hat{\y} = \q(\x, \h) = \vsig_y(\W^y_\out ((\W^x_\inp \x) \otimes (\W^h_\inp \h))).
\end{equation}

Equations (\ref{eq:hxy}) to (\ref{eq:yxh}) are identical to (\ref{eq:alg_yxh}) and (\ref{eq:alg_xh}), and thus they implement (\ref{eq:factors}), provided that the following weight tying rules are used\footnote{Different papers choose different conventions for deciding which matrix is the original and which is the transposed, see for instance \cite{im_GAE}, giving rise to different equations to implement (\ref{eq:hxy}) to (\ref{eq:yxh}).}: $\W^x=\W^x_\inp = {\W^x_\out}^\intercal$, $\W^y=\W^y_\inp = {\W^y_\out}^\intercal$ and $\W^h=\W^h_\inp = {\W^h_\out}^\intercal$. A benefit of using such weight tying rules is that it further reduces the number of parameters. Besides, any pair of the sub-networks described in (\ref{eq:hxy}) to (\ref{eq:yxh}) shares just one input matrix.

From the above presentation, it should be clear that the standard gated network architecture is completely symmetric: the functional role of the $\x$, $\y$ and $\h$ layers can be exchanged without changing the computations.

\subsection{Variations on the central tensor}

The architecture outlined in Section~\ref{sec:constrain} can be seen either as a particular way to parametrize the global 3-way tensor, introducing features into its internal structure, or as a way to replace the central tensor of the approach outlined in Section~\ref{sec:project} by an element-wise product of factor layers. This approach to implementing the central 3-way tensor can be seen as a degenerate case where all its non-diagonal elements are null and its diagonal elements are all set to 1. With this definition, the central product does not contain any tunable parameters. Instead, representation learning is implemented by tuning the weights of the $\W^x$, $\W^y$ and $\W^h$ matrices connecting the external layers to the factors. Note that using parameters instead of ones onto the diagonal may increase the flexibility of the model for learning, but it would not improve its expressive power, since the effect of changing these parameters can be captured by changing the parameters of the $\W$ matrices.

\begin{figure}[htb]
\centering
\includegraphics[width=0.8\textwidth]{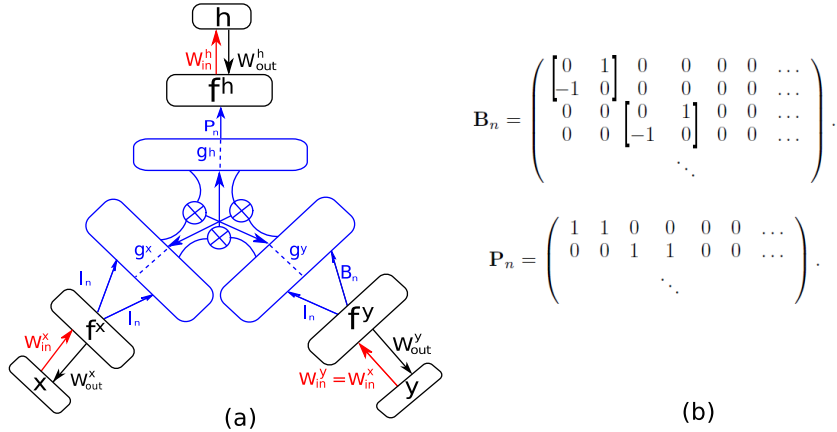}
\caption{\label{fig:gated_droniou} (a) In this architecture, an element-wise product is performed between vectors $\g^x$, $\g^y$ and $\g^h$ of size $2n$, where $n$ is the size of $\f^x$, $\f^y$ and $\f^h$.
The vector $\g^x$ is obtained by duplicating $\f^x$, using twice the identity matrix. One half of $\g^y$ is identical to $\f^y$, the other half is obtained through the block diagonal $\B_n$ matrix shown in (b). Finally, $\g^h$ is obtained from $\f^h$ by applying the $\MP_n$ matrix. Note that the weight-tying rules differ from the ones of standard gated networks.
}
\end{figure}

The constraint given in (\ref{eq:factors}) is somewhat arbitrary. For instance, the central computation of a gated architecture can be more complex than a simple element-wise product of factors. The architecture proposed in \cite{droniou2013gated} is an instance of such more complex computation. 
As outlined  in \figurename~\ref{fig:gated_droniou}, it also uses factors and a parameter-free tensor, but the structure of the central tensor has been specifically designed to learn orthogonal transformations. Several motivations for performing the corresponding computations are given in \cite{droniou2013gated}, together with the detailed mathematical rationale for such computations.

Note also that, in this architecture, the weight tying rules are unusual. Instead of having $\W_\inp = {\W_\out}^\intercal$ for all factors, $\W^h_\inp$ and $\W^h_\out$ are untied and $\W^x_\inp = \W^y_\inp$, with standard input-output weight tying rules on the $\x$ and $\y$ layers, i.e. $\W^x=\W^x_\inp = {\W^x_\out}^\intercal$ and $\W^y=\W^y_\inp = {\W^y_\out}^\intercal$.
A consequence of this choice is that the model might not be interpreted as an energy-based dynamical system, since $\W^h_\inp = {\W^h_\out}^\intercal$ is required so that Poincar\'{e}'s integrability criterion holds \cite{im_GAE}.

\subsection{Activation functions}
\label{sec:activ}

The $\x$, $\y$ and $\h$ layers can be either binary of real-valued. Depending on this format, different non-linear {\em activation} functions are used, resulting in different functionalities assigned to the network.

When the content of a size $n$ layer is binary, there are two options.
First, it can represent $2^n$ elements of a discrete set using standard binary coding rules. In that case, either the model directly represents the probability of each binary value, as is the case in \rbms, or the binary values are obtained from real-valued numbers by using a threshold. This latter case is uncommon because of the non-differentiability of the threshold function. The second option is to represent only $n$ elements using a ``one-hot'' representation where one bit is set to 1 and all the others are 0. When the corresponding layer is used as input, this representation is easy to enforce from the external world. 

For real-valued layers, the role of the activation function is to constrain the values of the output layer into a bounded domain, which can be obtained for instance with a sigmoid or a rectified linear unit, the latter being more popular in large network due to its faster computation. One can also use the {\em softplus} function $\sigma_+$, defined as $\sigma_+(x) = \log(1+\exp(x))$, which is a smooth version of the rectified linear unit \cite{glorot2011deep}.

To get a representation that is close toa ``one-hot'' in a real-valued output layer, the activity of the most active neuron in that layer can be highlighted by using the {\em softmax} function. If used for instance on the $\h$ layer of a gated network, the {\em softmax} function is

\begin{equation}\label{eq:softmax}\h = \vsig_{max}(\W^h_\out \mathbf{\f_h}),\end{equation}
where
\begin{equation}\sigma^i_{max}(\h=(h_1,\dots,h_n)) = \frac{e^{h_i}}{\sum_j e^{h_j}}.\end{equation}

In addition to highlighting the most active neuron(s), this function makes sure that all activities sum to 1. Hereafter, we call the obtained representation a ``soft one-hot''. 

Finally, if a binary ``one-hot'' representation was required as output, one could apply a postprocessing ``winner-takes-all'' function to a {\em softmax} layer, but we are not aware of any such use.

\section{Learning in gated networks}
\label{sec:learn}

Gated networks have two input layers and one output layer. One way to train such networks would be to use {\em supervised learning}: for a given pair of input layers, one would provide the expected output, and then train the network to minimize a function of the error between the expected and the obtained output. This is used in gated \cnns and gated \rnns (see Section~\ref{sec:beyond}).
But \grbms and \gaes are not trained in this way. Instead, the training process is designed to perform {\em unsupervised} learning, but differs between \grbms and \gaes. 
In this paper, we do not cover training \grbms, which is based on training \rbms. We refer the reader to \cite{swersky2010tutorial} for a clear presentation of the latter topic. Instead, we focus on training \gaes.

Given two input layers, \gaes are trained to {\em reconstruct} one of them.
In order to explain this learning process, it is useful to recap how it is performed in autoencoders.

An autoencoder is composed of two functions:
\begin{itemize}
\item
The encoding function that transforms the input vector $\x$ into a latent representation $\h$.
A typical function is $\h = \enc(\x) = \vsig_h(\W \x + \bb)$.
\item
The decoding function that reconstructs a representation $\hat{\x}$ of $\x$ from its latent representation $\h$.
A typical function is $\hat{\x} = \dec(\h) = \vsig_x(\W' \h + \bb')$.
\end{itemize}

The cost function for autoencoders is generally related to the reconstruction error. This error is for instance the distance between $\x$ and $\hat{\x}$, typically the squared error $||\hat{\x}-\x||^2$.
Learning then corresponds to applying an optimization algorithm such as a gradient-descent to the weights of the network so as to minimize this cost function.  Thus, during training, the network learns the encoding function and the decoding function simultaneously, using
$\hat{\x} = \vsig_x(\W' \vsig_h(\W \x + \bb) + \bb')$.
The main outcome of this learning process is the generation of the latent representation $\h$, that must be informative enough about the input so as to allow its correct reconstruction.


To highlight the relationship between autoencoders and \gaes, we now consider that $\h$ is the latent representation and $\x$ and $\y$ are the input layers.
%
%
Recalling (\ref{eq:hxy}) to (\ref{eq:yxh}), there are two ways to define a \gae as equivalent to an autoencoder.
The encoding function is always $\h = \oo(\x, \y)$, while the decoding function can be either

\begin{equation}
\label{eq:rx}
\hat{\x} = \p(\y, \h) = \p(\y, \oo(\x, \y))
\end{equation}

or

\begin{equation}
\label{eq:ry}
\hat{\y} = \q(\x, \h) = \q(\x, \oo(\x, \y)).
\end{equation}

Using (\ref{eq:hxy}) to (\ref{eq:yxh}), \eqref{eq:rx} can be rewritten
\begin{equation}
\label{eq:decx}
\hat{\x} = \vsig_x(\W^x_\out ((\W^y_\inp \y) \otimes (\W^h_\inp \vsig_h(\W^h_\out((\W^x_\inp \x) \otimes (\W^y_\inp \y)))))),
\end{equation}

and \eqref{eq:ry} can be rewritten
\begin{equation}
\label{eq:decy}
\hat{\y} = \vsig_y(\W^y_\out ((\W^x_\inp \x) \otimes (\W^h_\inp \vsig_h(\W^h_\out((\W^x_\inp \x) \otimes (\W^y_\inp \y)))))).
\end{equation}

One can note that $\W^y_\out$ does not appear in \eqref{eq:decx} and $\W^x_\out$ does not appear in \eqref{eq:decy}, thus tuning
these weights is not useful during training unless adequate weight tying rules are applied.


As outlined in Section~\ref{sec:history}, autoencoders can be endowed with properties similar to those of \rbms by using a denoising regularization function. There are three kinds of such functions, namely Gaussian noise, masking noise and salt and pepper noise \cite{rudy2014generative}.
It is commonplace to apply to \gaes these regularization functions as they are to autoencoders.
They are generally applied to all factor layers, but there are some exceptions. For instance, in \cite{rudy2014generative},
the denoising function is applied to $\x$ only.

Importantly, minimizing the squared reconstruction error of a \dae implements a regularized form of {\em score matching} \cite{vincent11}, which is itself a training criterion that favors the encoding of the manifolds where most of the input data is lying \cite{hyvarinen05}. 
The same applies to \gaes, but the nature of the represented manifolds depends on the encoded input-output relationships and on the format of the external layers.
Besides, some other regularizations functions such as {\em dropout} \cite{srivastava2014dropout} might also be applied to \gaes, but we are not aware of any work in this direction. For other practical hints on training gated networks, see also \cite{memisevic2013learning}.

Finally, the back-propagation algorithm can perform gradient descent on the weights of some or all the implied $\W$ matrices.

Taken together, the reconstruction function, the regularization function and the learning rules define many different settings to learn representations with \gaes.
We study other combinatorial aspects in the next section.

\section{Applications of gated networks}
\label{sec:views}

Given what we have presented so far, there are three respects in which the use of gated networks may vary.
First, as outlined in Section~\ref{sec:activ} the content of the $\x$, $\y$ and $\h$ layers is either binary, one-hot or real-valued.
Second, as outlined in Section~\ref{sec:learn}, gated networks can be trained in various ways using various training signals, regularization functions and cost functions.
Third, different layers can be used either as input or output.
All these variations give rise to different functional roles for the corresponding networks.
The goal of this section is to make an inventory of such uses in the literature, which is finally summarized in Table~\ref{tab:inventory}.

\subsection{Format of the external layers}

In Section~\ref{sec:activ}, we outlined the different activation functions that are used to deal with different format of the external layers.
Here, we recapitulate the use of these formats in different models.

First, in all \grbms, the $\h$ layer always uses standard binary encoding \cite{Memisevic2007,Memisevic2010}.

Furthermore, most models use pixels of two images as $\x$ and $\y$ input. The transformation between these images stored in $\h$ is either binary \cite{Memisevic2007,Memisevic2010} or real-valued \cite{droniou2013gated,dehghan2014look}.
In both cases, what is learned is a manifold of the pixels in the $\x$ conditioned on those of the $\y$ layer, or {\em vice versa} \cite{Memisevic2007}.

There are two models where the $\y$ layer is binary. First, the {\em gated softmax classification} model was used in the context of logistic regression, i.e. classification using a log-linear model, where the output $\hat{\y}$ consisted of binary class labels,  and the values of the $\h$ layer were also binary \cite{Memisevic2010b}. 

More recently, in the context of studying the generative property of \gaes, a model was proposed where the $\y$ input also consists of a class-conditional, one-hot representation, whereas $\h$ is a real-valued representation constrained by a rectified linear unit \cite{rudy2014generative}. The network is trained to regenerate examples from the MNIST and Toronto Faces Database images, thus $\x$ is a vector of pixels. In this context, the model represents class-conditional manifolds, i.e. a set of manifolds of the input data $\x$ with one manifold per corresponding class in $\y$. As the authors state, this use of the \gae ``is akin to learning a separate \dae model for each class, but with significant weight sharing between the models. In this light, the gating acts as a means of modulating the model's weights depending on the class label'' \cite{rudy2014generative}.

\subsection{Training signal}
\label{sec:training_signal}

As outlined in Section~\ref{sec:learn}, \gaes can be trained to reconstruct either $\hat{\x}$ or $\hat{\y}$. When the input data is binary, the cross-entropy loss function is the default choice \cite{rudy2014generative}. When it is real-valued, the standard cost function is a squared reconstruction error.
Therefore, when training to reconstruct $\hat{\x}$, it is $J = \frac{1}{2} || (\hat{\x}|\y) - \x ||^2$, whereas for reconstructing $\hat{\y}$, it is $J = \frac{1}{2} || (\hat{\y}|\x) - \y ||^2$.

The first option is the one chosen in \cite{rudy2014generative}. 
This makes the connection to autoencoders more explicit because they both take $\x$ as input and $\hat{\x}$ as output. But this contrasts with the rest of the literature, where it is more common to train to reconstruct $\hat{\y}$ \cite{Memisevic2007,Memisevic2010,Memisevic2010b,droniou2013gated,michalski2014modeling,michalski2014b}.

A third option exists. If we want the model to be able to reconstruct $\hat{\x}$ given $\y$ and $\hat{\y}$ given $\x$ at the same time, we can use \cite{memisevic2011gradient}:
\begin{equation}
\label{eq:sym}
J = \frac{1}{2} || (\hat{\x}|\y) - \x ||^2 + \frac{1}{2} || (\hat{\y}|\x) - \y ||^2.
\end{equation}

A particularly relevant case for using this symmetric signal is the case where $\x = \y$. In that case, the mapping units $\h$ learn covariances within $\x$ \cite{memisevic2011gradient}.



Interestingly, a model recognizing offspring relationship from pictures of faces combines generative and discriminative training, using two training signals  \cite{dehghan2014look}. From one side, it learns a representation of the transformation between two faces using the symmetric cost function given in \eqref{eq:sym}. But it also tries to determine offspring relationship as a binary representation, so it uses a softmax cost function during a supervised label learning process. Finally, both cost functions are combined into a weighted sum.

\subsection{Input-Output function}

We have outlined in Section~\ref{sec:constrain} that the role of the $\x$, $\y$ and $\h$ layers could be exchanged.
This leads to three permutations where two layers among $\x$, $\y$ and $\h$ are inputs, the third layer being the output.
However, given the unsupervised training procedure described in Section~\ref{sec:learn}, we see that, in addition to the three possibilities outlined above, one can also use it to predict either $\hat{\x}$ or $\hat{\y}$. Under this view, learning the latent representation $\h$ is a side effect, $\h$ being used neither as input nor as output, but being ``reinjected'' into the network to reconstruct one of the input layers. The same fact applies {\em mutatis mutandis} to all other layers.

The different possible output layers result in two main ways to use a gated network.
The first one, the predictive coding view, consists in inferring an output $\hat{\y}$ (or $\hat{\x}$) given an input $\x$ and a context $\h$.
The {\em temporal} predictive coding view is a special case of the above, with $\x_t$ as input and $\x_{t+1}$ as output.
The second one, the {\em transformation coding view}, consists in using the latent representation $\h$ as output, given two input vectors $\x$ and $\y$.
The output layer $\h$ then expresses some relations between $\x$ and $\y$, which may provide abstract representations that can be used for instance in higher level decision modules. 

The latter view is mostly used to learn transformations between two successive images, so as to extract features containing temporal information \cite{Memisevic2007,Memisevic2010,Memisevic2010b,droniou2013gated,michalski2014modeling,michalski2014b}. In this context, the input vectors $\x$ and $\y$ are successive images, for instance from a video.
The extracted transformations $\h$ are content-independent. For instance, they can represent rotations, independently from what is rotated in the images.
Furthermore, they  convey a temporal information about these successive images, thus they can be used as elementary features in a higher level to model some temporal information. However, we are not aware of any architecture where these temporal features are actually used to extract  temporally extended information from videos, apart from very preliminary attempts among 3 or 4 successive frames in \cite{michalski2014modeling} using a hierarchical sequence of \gaes (see Section~\ref{sec:beyond}).
The work of \cite{dehghan2014look} is another instance of the transformation coding view, but where $\x$ and $\y$ are temporally independent images.

In many papers, both the transformation representation $\h$ and the reconstructed input signal $\hat{\x}$ or $\hat{\y}$ are studied.
As a consequence, in the absence of an external architecture that uses it, it is often hard to determine which of these signals should be considered as the output of the network. Moreover, it is often the case that, when learning transformations between two successive images, the learned transformation is then applied to a new input image to see what output image is ``fantasized'' by the network, performing a type of ``analogy making''. In this context, the output of the network is both $\hat{\h}$ and $\hat{\y}$ \cite{Memisevic2007,Memisevic2010,Memisevic2010b,droniou2013gated,michalski2014modeling,michalski2014b}.
Thus in Table~\ref{tab:inventory}, we do not strive to determine which layer is the output of the studied algorithm.

\subsubsection{Summary: an inventory}

Table~\ref{tab:inventory} summarizes many uses of the standard gated networks listed above.

\begin{table}[hbtp]
\begin{center}
\begin{tabular}{|c||c|c|c|c|c|}
    \hline
Papers & $\x$ & $\y$ & $\h$ & act. func. & training \tabularnewline
    \hline 
\cite{Memisevic2007,Memisevic2010} & pixels(t) & pixels(t+1) & binary  & proba & $\hat{\y}$  \tabularnewline
    \hline
\cite{Memisevic2010b} & pixels & binary & binary  & proba & $\hat{\y}$ \tabularnewline
    \hline
\cite{memisevic2011gradient} & pixels & pixels = $\x$ & soft 1-hot & relu & $(\hat{\x},\hat{\y})$ \tabularnewline
    \hline
\cite{droniou2013gated} & pixels(t) & pixels(t+1) & real & softplus & $\hat{\y}$ \tabularnewline
    \hline
\cite{rudy2014generative} & pixels & 1-hot & real & relu & $\hat{\x}$  \tabularnewline
    \hline 
\cite{dehghan2014look} & face 1 & face 2 & soft 1-hot & softmax & {\em hybrid} \tabularnewline
    \hline 
\end{tabular}
\caption{Various input-output functions for gated networks. ``act. func'' stands for the activation function on the $\h$ layer. ``relu'' stands for rectified linear unit, ``real'' stands for real-valued. ``proba'' stands for a probabilistic activation function. The $(\hat{\x},\hat{\y})$ training signal stands for the symmetric cost function given in \eqref{eq:sym}. 
For the {\em hybrid} training signal, see Section~\ref{sec:training_signal}.
\label{tab:inventory}}
\end{center}
\end{table}

Table~\ref{tab:inventory} illustrates that there is a wide variety of ways to use gated networks. This variety is even greater if we also consider the non-standard architectures surveyed in the next section.

\section{Beyond standard gated architectures}
\label{sec:beyond}

In this section, we describe a few architectures that contain a gated network. First, we list some architectures where the central tensor connects more than 3 layers. Then, we present some architectures whose set of connections is not restricted to the central tensor.

\subsection{Extended tensors}

\begin{figure}[hbtp]
\begin{center}
\includegraphics[width=0.5\textwidth]{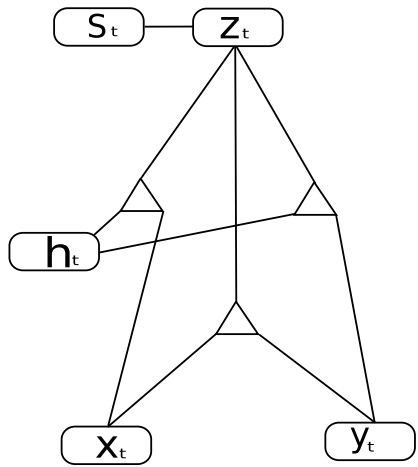}
\caption{Four layers can be connected by three tripartite connection blocks (adapted from \cite{Taylor2011}).
\label{fig:taylor}
}
\end{center}
\end{figure}

There are some architectures where the central tensor connects more than 3 external layers. Conditional \rbms (\crbms) are \rbms where some memory of the past input are included into the input layer so that the architecture can model time-dependent data \cite{Taylor2009}. In \cite{Taylor2011}, a \crbm is used to model human motion data but, as illustrated in \figurename~\ref{fig:taylor}, it is extended with an additional {\em style} layer to model different styles of motion.

The $\x$ layer corresponds to the motion input at previous time step. The $\y$ layer, which is the output, corresponds to the predicted motion at the current time step.
The $\h$ layer is used as in all \grbms to learn a representation of the transformation between $\x$ and $\y$. But, additionally, the $\z$ layer corresponds to real-valued stylistic features that are fed by discrete {\em style} labels (encoded in the $\style$ layer) and provide some additional {\em contextual} information about the motion.

The resulting architecture is then factored as described above so as to limit the number of parameters, but it is designed in such a way that the 4 factors are only connected together by triplets using factored 3-way tensors.

More recently, a ``4-way tensor'' and its factorization were introduced based in \grbms \cite{mocanu2015factored}. The central factored operation consists in performing a sum of products of second order tensors. The models of \cite{Taylor2009} and \cite{mocanu2015factored} are both capable of representing sequential data in the limit of the $N$ previous time steps included in the memory concatenated to the input layer.




\subsection{Clustering with gated networks}

\begin{figure}[hbtp]
\begin{center}
\includegraphics[width=0.5\textwidth]{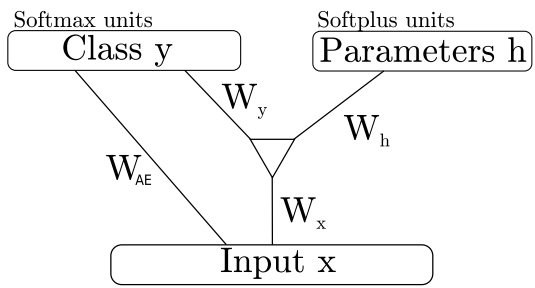}
\caption{Gated network for unsupervised classification (adapted from \cite{droniou14RAS}). The \gae is represented using the simplified notation of \figurename~\ref{fig:gated_connections}(c). With respect to a standard \gae, it uses an additional autoencoder implemented through the $\W_{AE}$ matrix.
\label{fig:droniouRAS}
}
\end{center}
\end{figure}

In some architectures, the central 3-way tensor is not the only ingredient. For instance, the architecture depicted in \figurename~\ref{fig:droniouRAS} uses an additional autoencoding connection with respect to a standard \gae \cite{droniou14RAS}.

The network aims at clustering input data into ``concepts'' corresponding to manifolds in the input layer, without using supervised learning. For doing so, the input $\x$ is first fed into a standard autoencoder using a softmax activation function that performs unsupervised clustering of the input data. The softmax activation function implements a competition between the bits of the class layer and results in the emergence of a soft one-hot representation of the corresponding class. Then, given the input and the obtained class, the $\h$ layer implements a parametrization of the input with respect to the class, using a softplus layer, i.e. $\h = \vsig_+ (\W^h_\out \mathbf{\f_h})$.

Since it uses a soft one-hot, class-conditional $\y$ layer and a real-valued input $\x$ layer, this model can be seen as a direct extension of the one presented in \cite{rudy2014generative}. However, since the weights are trained simultaneously, the network in \figurename~\ref{fig:droniouRAS} finds the adequate classes to represent the data with an accurate parametrization by itself, instead of requiring them as training labels. This endows the network with unsupervised clustering capabilities that are well beyond those of standard dimensionality reduction techniques. This model is then extended to deal with multimodal information, showing an even improved clustering performance. We do not further study this aspect here, see \cite{droniou14RAS} for more details.

\subsection{Recurrent gated networks}

\begin{figure}[hbtp]
\begin{center}
\includegraphics[width=0.3\textwidth]{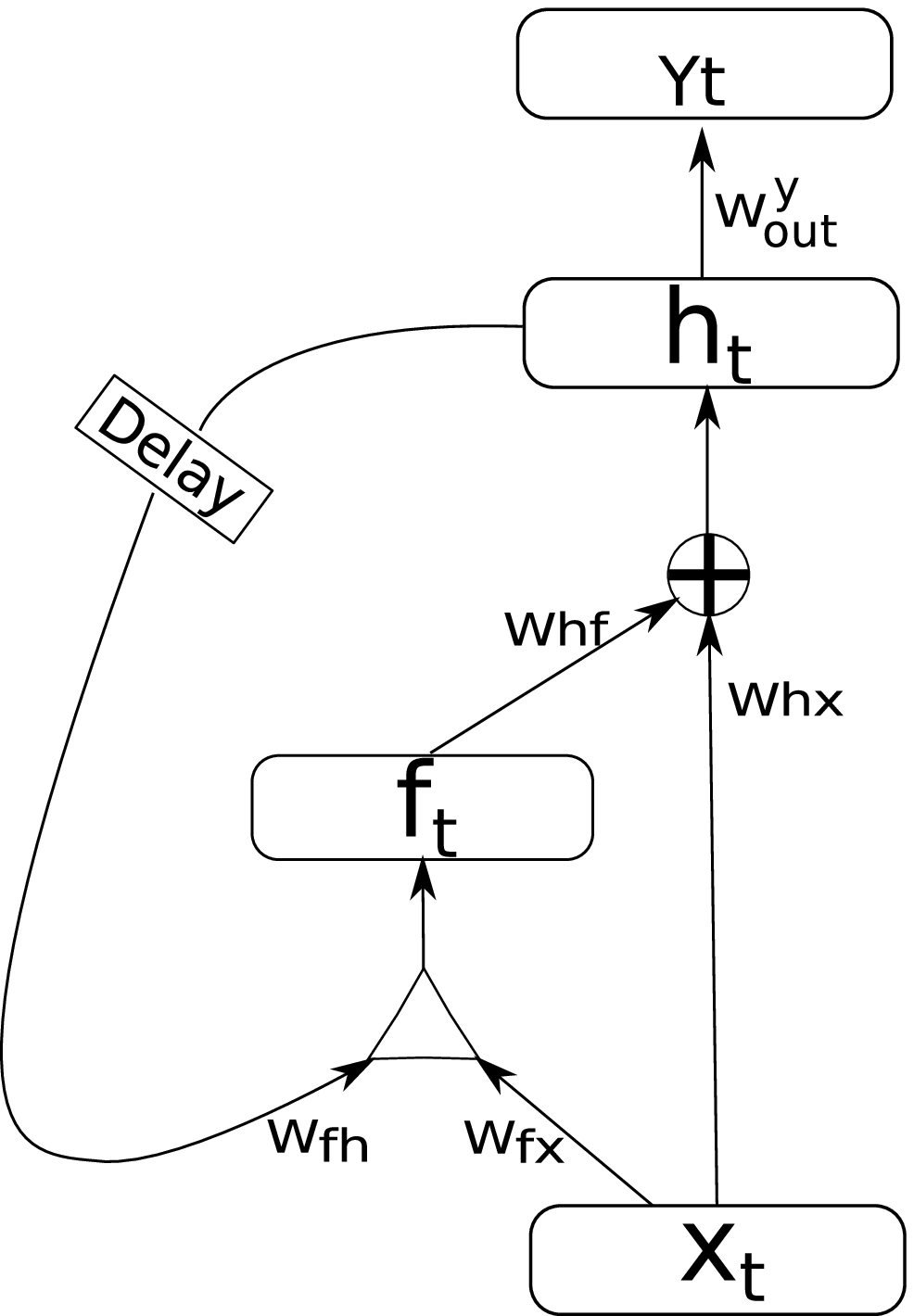}
\caption{Multiplicative \rnn.
\label{fig:sutskever}
}
\end{center}
\end{figure}

Another architecture based on factoring gating connections is the ``Multiplicative \rnn'' architecture \cite{sutskever2011generating} depicted in \figurename~\ref{fig:sutskever}. This is a recurrent architecture trained to deal with temporally organized information such a text or speech signal.

The key requirement of the architecture is that the recurrent connection responsible for the dynamics of the hidden variable should be a function of the input layer $\x$. This would lead to a full 3-way tensor, which the authors factorize as described in Section~\ref{sec:trick} to reduce the number of free parameters. With slightly adapted notations to highlight the similarity with other architectures, the internal computation of the multiplicative \rnn is given by the following equations:

\begin{align}
f_t &= \diag(\W_{fx} \x_t).\W_{fh} \h_{t-1} \label{eq:sutskever1}\\
\h_t &= \tanh(\W_{hf} f_t + W_{hx} \x_t) \label{eq:sutskever2}\\
\hat{\y}_t &= \W^y_\out \h_t + \bb_y. \label{eq:sutskever3}
\end{align}

A key difference between this work and the other ones presented above is that the architecture is trained in a supervised way, rather than trained to reconstruct its input. The focus is thus not on extracting an abstract representation of the input. Another originality is that, instead of being trained with a standard first order gradient descent algorithm such as back-propagation, the architecture is trained using a second order method based on Hessian-free optimization \cite{martens2010deep}. To our knowledge, despite its efficiency, no other gated network has been trained with this method.

\subsection{Convolutional gated networks}

Convolution is a technique which consists in processing a large image by shifting a smaller filter to any position in the image and applying it over all positions. For instance, the same filter can be applied to recognize a pattern at any position in the image. Convolutional gated networks apply the convolution idea to gated networks. This has been done in \grbms \cite{Taylor2010} so as to extract spatio-temporal features in the context of human activity recognition, and in \gaes \cite{konda2015learning} to perform visual odometry from stereo pairs in a sequence of images captured from a moving camera.

\subsection{Prediction with a sequence of gated networks}

Another architecture models temporal data using a sequence of \gaes \cite{michalski2014modeling}\footnote{\cite{michalski2014b} is the corresponding arXiv preprint}.
Beyond a sequence, it even uses a hierarchy of \gaes to learn transformations of transformations.
The model of \cite{michalski2014modeling}, called  Predictive Gating Pyramides (\pgp), cascades two level of \gaes to predict sequences. As the authors state, the reconstruction error is inadequate in their context, thus the model is trained explictly to predict rather than to reconstruct. Actually, it is trained to predict over multiple steps.
A strong assumption in \pgp is that the highest-order relational structure in the sequence is constant. It uses Back-Propagation Through Time (\bptt) to perform gradient descent on the weights over time. However, the model is used to learn temporal features, it does not predict long sequences of images.  And a major drawback is that the architecture requires as many \gaes as time steps.

\section{Conclusion}
\label{sec:conclusion}

In this paper, we have based our presentation of gated networks on a perspective that insists on their symmetric nature.
Based on this particular perspective, we could highlight its richness by providing an inventory of the various ways they have been used so far in the literature. Given this richness, we believe standard gated networks still have a largely underexploited potential as a unifying tool for many domains where the relevant information is naturally expressed as tripartite relationships between three interdependent sources. Apart from the ones proposed in this paper, we hope many other application domains to gated networks will emerge in the next years.

Furthermore, as pointed out in Section~\ref{sec:beyond}, there are still rather few non-standard architectures based on the factored gating idea. We believe the list of such architectures will expand in the future, and also that gated networks should be included into more general frameworks that may contain several instances of such networks, as is already the case with \cite{michalski2014modeling} or \cite{droniou14ICDL}.

Finally, among other things, an interesting perspective to this work consists in combining it with the {\em contextual learning} perspective  \cite{jonschkowski2015contextual}.
Indeed, several contextual learning patterns might be implemented with gated networks, and some works about representation learning with gated networks
might be interpreted in the framework of contextual learning.

\section*{Acknowledgments}
This work was supported by the European Union's Horizon 2020 research and innovation program within the DREAM project under grant agreement $N^o$ 640891.



\bibliographystyle{arxiv}
\bibliography{bibgae}
\end{document}